\pdfoutput=1
\documentclass[runningheads]{llncs}
\usepackage{graphicx}

\usepackage{comment}
\usepackage[ruled]{algorithm2e}
\usepackage{amsmath,amssymb} 
\usepackage{color}
\usepackage{subfigure}
\usepackage{wrapfig}
\usepackage{multirow}
\usepackage{tabularx}

\begin{document}
\pagestyle{headings}
\mainmatter
\def\ECCVSubNumber{8}  

\title{QuantNet: Learning to Quantize by Learning within Fully Differentiable Framework} 

\titlerunning{QuantNet}
%
\author{Junjie Liu\inst{1}\and
Dongchao Wen\inst{1}\orcidID{0000-0001-7311-1842} \and
Deyu Wang\inst{1}\and
Wei Tao\inst{1}\and
Tse-Wei Chen\inst{2}\and
Kinya Osa\inst{2}\and
Masami Kato\inst{2}}
\authorrunning{Junjie Liu et al.}
%
\institute{Canon Information Technology (Beijing) Co., LTD, China \\
\email{\{liujunjie, wendongchao, wangdeyu, taowei\}@canon-ib.com.cn}\\
\and
Device Technology Development Headquarters, Canon Inc., Japan\\
\email{twchen@ieee.org}}
\maketitle

\begin{abstract}
Despite the achievements of recent binarization methods on
reducing the performance degradation of Binary Neural Networks (BNNs),
gradient mismatching caused by the Straight-Through-Estimator (STE) still dominates quantized networks.
This paper proposes a meta-based quantizer named QuantNet,
which utilizes a differentiable sub-network to directly binarize the full-precision weights without resorting to STE and any learnable gradient estimators.
Our method not only solves the problem of gradient mismatching,
but also reduces the impact of discretization errors, caused by the binarizing operation in the deployment, on performance.
Generally, the proposed algorithm is implemented within a fully differentiable framework,
and is easily extended to the general network quantization with any bits.
The quantitative experiments on CIFAR-100 and ImageNet demonstrate that
QuantNet achieves the significant improvements comparing with previous binarization methods,
and even bridges gaps of accuracies between binarized models and full-precision models.
\end{abstract}

\section{Introduction}
Deep neural networks (DNNs) have achieved remarkable success in several fields in recent years.
In particular, convolutional neural networks (CNNs) have shown state-of-the-art
performance in various computer vision tasks such as image classification, object detection, trajectory tracking, etc.
However, an increasing number of parameters in these networks also
lead to the larger model size and higher computation cost,
which gradually becomes great hurdles for many applications,
especially on some resource-constrained devices with limited memory space
and low computation ability.

To reduce the model size of DNNs,
representative techniques such as
network quantization \cite{Zhou2016DoReFa,hou2016loss,pact2018,jung2019learning,lahoud2019self},
filters pruning \cite{Luo2017ThiNet,he2017channel,liu2017learning},
knowledge distillation \cite{Hinton2015Distilling,BSS2019,liu2019kr}
and deliberate architecture design \cite{Li2017,cheng2017survey} are proposed.
As one of typical solutions, the quantization based method quantizes
floating-point values into discrete values in order to generate the quantized neural networks (QNNs) as compact as possible.
In the most extreme case,
if both network weights and network activations are binarized (BNNs) \cite{courbariaux2015binaryconnect},
the computation can be efficiently implemented via bitwise operations,
which enables about 32 $\times$ memory saving and 58 $\times$ speeding up \cite{rastegari2016xnor} on CPUs in inference.

Despite the advantages we mentioned above,
how to alleviate performance degradation of quantized networks is still under research,
especially for binarized networks.
In general, BNNs involve a $sign$ function to obtain signs of parameters.
The non-differentiable sign function leads to gradient vanishing almost anywhere.
To address this issue, some works \cite{liu2020bamsprob,Zhu2020} propose low-bit training algorithms to relieve the impact of gradients quantization errors,
and another works focus on estimating the vanishing gradients. The Straight-Through Estimator (STE) \cite{ste2013}
is commonly used to estimate the vanishing gradients during the back-propagation,
while the well-known gradient mismatching problem \cite{yin2019understanding,hou2019analysis,binaryop2019} is introduced.

As the number of quantized bits decrease,
the gradients estimated by STE depart further from the real gradients.
Thus, the gradient mismatching is considered as the main bottleneck of performance improvements of binarized models.
As one of promising solutions, estimating more accurate gradients is suggested by recent methods.
Some of these methods \cite{Liu2018Enhance,Wang2018Twostep,blended_grad2018,Uhlich2019DQ}
try to refine the gradients estimated by STE with extra parameters,
and others \cite{bai2019prox,chen2019meta} address the problem
by replacing STE with learnable gradient estimators.
Different from these efforts on estimating more accurate gradients,
the individual method \cite{lahoud2019self} employs a differentiable function $tanh$ as a soft binarizing operation,
in order to replace the non-differentiable function $sign$.
\emph{Thus, it will no longer require STE to estimate gradients}.

In this paper, we follow the idea of soft binarization,
but we focus on solving two important issues that are left out.
Firstly, although the soft binarization solves the problem of gradient mismatching,
another issue of gradient vanishing from the function $tanh$ arises.
It not only causes the less ideal convergence behavior, but makes the solution highly suboptimal.
Moreover, as the soft binarization involves a post-processing step,
how to reduce the impact of the discretization errors on performance is very important.
With these motivations, \emph{we propose a meta-based quantizer named QuantNet
for directly generating binarized weights with an additional neural network.
The said network is referred to as a meta-based quantizer and optimized with the binarized model jointly.}
In details, it not only generates the higher dimensional manifolds of weights for easily finding the global optimal solution,
but also penalizes the binarized weights into sparse values with a task-driven priority for minimizing the discretization error.
For demonstrating the effectiveness of our claims,
we present the mathematical definition of two basic hypotheses in our binarization method,
and design a joint optimization scheme for the QuantNet.

We evaluate the performance of our proposed QuantNet by comparing it with the existing binarization methods
on the standard benchmarks of classification task with
CIFAR-100 \cite{cifar10} and ImageNet \cite{Deng2009}.
As for the baseline with different network architectures,
AlexNet \cite{Krizhevsky2012}, ResNet \cite{he2016deep},
MobileNet \cite{sandler2018mobilenetv2} and DenseNet \cite{denseNet2017} are validated.
The extensive experiments demonstrate that our method achieves remarkable improvements
than state-of-the-arts across various datasets and network architectures.

In the following,
we briefly review previous works related to network quantization in section 2.
In section 3, we define the notations and present the mathematical definition of existing binarization methods.
For section 4, we present two basic hypotheses of our binarization method and exhibit the implementation details.
Finally, we demonstrate the effectiveness and efficiency of our method in section 5,
and make the conclusions in section 6.


\section{Related Work}
In this section, we briefly review existing methods on neural network quantization.
As the most typical strategy to achieve the purpose of network compression,
the network quantization has two major benefits - reducing the model size while improving the inference efficiency.
Comparing with the strategies of network filters pruning \cite{Luo2017ThiNet,he2017channel,liu2017learning}
and compact architecture design \cite{Li2017,cheng2017survey},
how to alleviate the performance degradation in quantized model \cite{hou2019analysis} is still unsolved,
especially for the binarized model \cite{yin2019understanding,alizadeh2019a,binaryop2019}.

\paragraph{\textbf{Deterministic Weight Quantization}}
Through introducing a deterministic function,
traditional methods quantize network weights (or activations) by minimizing quantization errors.
For examples, BinaryConnect \cite{courbariaux2015binaryconnect} uses a stochastic function
for binarizing weights to the binary set \{+1,-1\},
which achieves better performance than two-step approachs \cite{Han2016compress,Kim2016compress} on several tasks.
Besides, XNOR-net \cite{rastegari2016xnor} scales the binarized weights
with extra scaling factors and obtains better results.
Furthermore, Half-Wave-Gaussian-Quantization (HWGQ) \cite{Cai2017Deep}
observes the distribution of activations,
and suggests some non-uniform quantization functions for constraining unbounded values of activations.
Instead of binarizing the model,
the ternary-connect network \cite{Zhu2016Ter} and DoReFa-Net \cite{Zhou2016DoReFa}
perform the quantization with multiple-bits via various functions to bound the range of parameters.

\emph{These methods purely focus on
minimizing quantization errors between full-precision weights and quantized weights,
however less quantization errors do not necessarily mean better performance of a quantized model.}

\paragraph{\textbf{Loss-Aware Weight Quantization}}
As less quantization errors do not necessarily mean better performance of a quantized model,
several recent works propose the loss-aware weight quantization in terms of
minimizing the task loss rather than quantization errors.
The loss-aware binarization (LAB) \cite{hou2016loss} proposes
a proximal Newton algorithm with diagonal Hessian approximation that
minimizes the loss with respect to the binarized weights during optimization.
Similar to LAB, LQ-Net \cite{Zhang2018LQ}
allows floating-point values to represent the basis of quantized values during the quantization.
Besides, PACT \cite{pact2018} and SYQ \cite{Faraone2018syq}
suggest parameterized functions to clip the weights or activation value during training.
In the latest works, QIL \cite{jung2019learning} parameterizes the non-linear quantization intervals and
obtains the optimal solution by minimizing with the constraint from task loss.
And self binarization \cite{lahoud2019self} employs a soft-binarization function
to evolve weights and activations during training to become binary.

\emph{In brief, through introducing learnable constraints or scaling factors,
these methods alleviate the performance degradation in their quantized models,
but the gradient mismatching problem caused by the $sign$ function and STE \cite{alizadeh2019a} is still unconsidered.}

\paragraph{\textbf{Meta-Based Weight Quantization}}
As the quantization operator in training process is non-differentiable,
which leads to either infinite gradients or zero gradients,
MixedQuant \cite{Uhlich2019DQ} addresses a gradient refiner by introducing the assistant variable
for approximating the more accurate gradients.
Similar to MixedQuant, ProxQuant \cite{bai2019prox} proposes an alternative approach that
formulates quantized network training as a regularized learning problem
and optimizes it by the proximal gradients.
Furthermore, Meta-Quant \cite{chen2019meta} proposes a gradient estimator to directly
learn the gradients of quantized weights by a neural network,
in order to remove STE commonly used in back-propagation.

\emph{Although such methods have noticed that refining the gradients computed by STE
or directly estimating the gradients by meta-learner is helpful to alleviate the problem of gradient mismatching,
the increasing complexity of learning gradients introduces a new bottleneck.}

\section{Preliminaries}
\paragraph{\textbf{Notations}}
For a vector $x$, where $x_i; i \leq n$ is the element of $x$,
we use $\sqrt{x}$ to denote the element-wise square root,
$|x|$ denotes the element-wise absolute value,
and $\|x\|_p$ is the $p$-norm of $x$.
$sign(x)$ is an element-wise function denoting that
$sign(x_i)=1; \forall i \leq n$ if $x_i \geq 0$ and -1 otherwise.
We use $diag(X)$ to return the diagonal elements of matrix $X$,
and $Diag(x)$ to generate a diagonal matrix with vector $x$.
For two vectors $x$ and $y$,
$x \odot y$ denotes the element-wise multiplication
and $x \oslash y$ denotes the element-wise division.
For a matrix $X$, $vec(X)$ denotes to return a vector by stacking all the columns of $X$.
In general, $\ell$ is used to denote the objective loss,
and both $\partial \ell / \partial x$ and $\nabla \ell (x)$ denote
the derivative of $\ell$ with respect to $x$.

\paragraph{\textbf{Background of Network Binarization}}
The main operation in network binarization is the linear (or non-linear) discretization.
Taking a multilayer perception (MLP) neural network as an example, one of its hidden layers can be expressed as

\begin{equation}\label{eq:binary_basic}
\textbf{w}_q = f(\textbf{w})^r \odot binarize(\textbf{w})
\end{equation}

where $\textbf{w} \in \mathbb{R}^{m \cdot n}$ is the full-precision weights,
and $m, n$ are the number of input filter channels, the number of output filter channels\footnote{In this paper,
the kernels on full connected layer are regarded as a special type of the convolutional kernels}, respectively.
Based on the full-precision (floating-point) weights, the corresponding binarized weights $\textbf{w}_q$
is computed by two separate functions $f(\textbf{w})^r$ and $binarize(\textbf{w})$,
and the goal is to represent the floating-point elements in $\textbf{w}$ with one bit.

In BinaryConnect \cite{courbariaux2015binaryconnect},
$f(\textbf{w})^r$ is defined as the constant 1,
and $binarize(\textbf{w})$ is defined by $sign(\textbf{w})$,
which means each element of $\textbf{w}$ will be binarized to \{-1,+1\}.
For XNOR-Net \cite{rastegari2016xnor},
it follows the definition of BinaryConnect on $binarize(\textbf{w})$,
but further defines $f(\textbf{w}_t)^r = \|\textbf{w}_t\|_1 / (m \times n)$
and $r$ is defined as the constant 1,
where $t$ is the current number of training iterations.
Different from the determining function on $f(\textbf{w})$,
the Loss-Aware Binarization (LAB) \cite{hou2016loss} suggests
a task-driven $f(\textbf{w}_t)$ with the definition of
$\|d_{t-1} \odot \textbf{w}_t\|_1 / \|d_{t-1}\|_1$,
where $d_{t-1}$ is a vector containing the diagonal $diag(D_{t-1})$ of
an approximate Hessian $D_{t-1}$ of the task loss.
Furthermore, QIL \cite{jung2019learning} extends $f(\textbf{w}_t)^{r_t}$
into a nonlinear projection by setting $r_t$ to be learnable and $r_t > 1$ for all $t$.
Considering the back-propagation,
as STE \cite{yin2019understanding} with $sign$ function
introduces the major performance bottleneck for BNNs,
Self-Binarization \cite{lahoud2019self} defines $binarize(\textbf{w})$ as $tanh(\textbf{w})$.
In the training of BNNs,
the $tanh$ function transforms the full-precision weights $\textbf{w}$
to obtain weights $\textbf{w}_q$ that are bounded in the range [-1,+1],
and these weights are closer to binary values as the training converges.
After the training, $\textbf{w}_q$ are very close to the exact set of \{+1,-1\},
and the fixed point values will be obtained by taking the sign of the $\textbf{w}_q$.

\section{Methodology}
In this section,
we firstly present the mathematical definition of two basic hypotheses in our binarization method.
Then we propose a meta-based quantizer named QuantNet for directly generating
binarized weights within a fully differentiable framework.
Moreover, a joint optimization scheme implemented in a standard neural network training is designed to solve the proposal.

\subsection{Assumptions}
As can be seen,
the work \cite{lahoud2019self} replaces the hard constraint $sign$ with the soft penalization $tanh$,
and penalizes the output of $tanh$ to be the closest binary values.
However, there are two important issues which are ignored.

In the case of binarizing weights with $tanh$,
as most of the elements in binarized weights are close to \{+1,-1\} at the early stage of training,
these elements will reach saturation simultaneously, and then cause the phenomenon of gradients vanishing.
In brief, if the element is saturated on +1, it will not be able to get close to -1 again.
On the contrary, the case of the element saturated on -1 is the same.
It means that flipping values of these elements is impossible.
As a result, only a few unsaturated elements will oscillate around zero,
which causes the less ideal convergence behavior and makes the solution highly suboptimal.

Moreover, different from the hard constraint methods,
the soft penalization method contains a post-processing step with rounding functionality,
and it rounds the binarized weights for further obtaining the fixed point (discrete) values.
With the increasing number of the network parameters,
the discretization error caused by the rounding function will be the major factor to limit performances.
To propose our method for solving above issues, we make two fundamental hypotheses in the following.

\begin{figure}[t]
  \centering
  \includegraphics[width=0.95\textwidth]{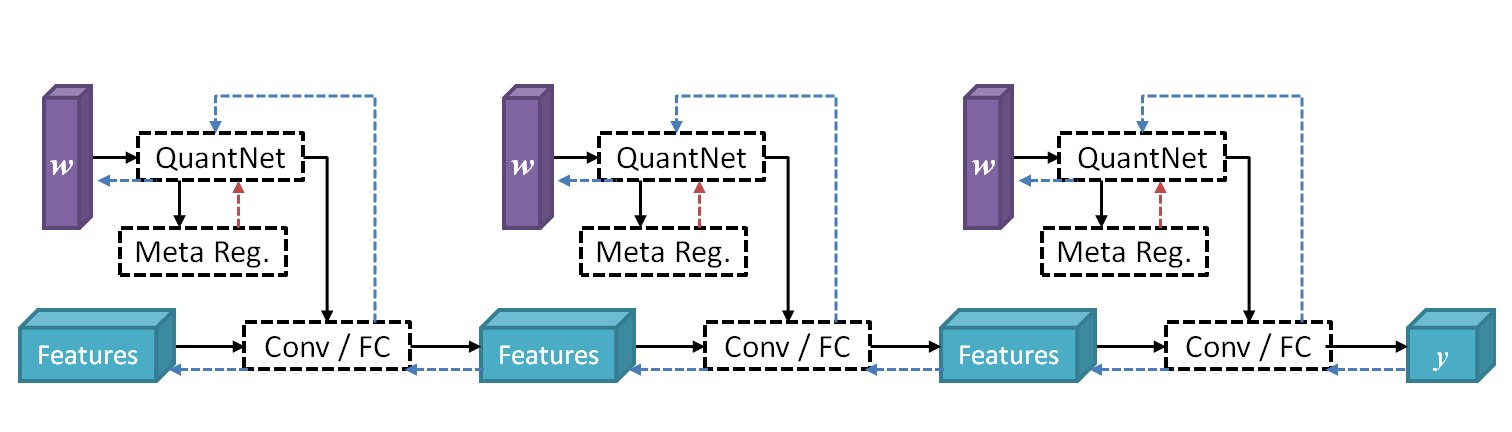}
  \caption{The architecture of our binarization method. The solid and dashed (blue color) lines represent the feed-forward process and the gradient flow of back-propagation separately, and the dashed line with red color means the meta-gradient flow from the meta regularization (Best viewed in color).}\label{fig:pipeline}
\end{figure}

\paragraph{\textbf{Assumption 1}:}
we assume that there exists the functional $\mathcal{F}$ to form $tanh(\mathcal{F}(\textbf{w}))$,
and $\lim \limits_{\textbf{w} \rightarrow \infty}
\nabla \mathcal{F}(\textbf{w}) \cdot (1 - tanh^2 ( \mathcal{F}(\textbf{w})) \neq 0$,
then the derivative of $tanh()$ with respect to $\textbf{w}$ is expressed as

\begin{equation}\label{eq:hypo1}
\begin{split}
&\lim \limits_{\textbf{w} \rightarrow \infty}
\frac{\partial tanh ( \mathcal{F}(\textbf{w}) )}{\partial \textbf{w}} \neq 0 \\
w.r.t. \quad & \nabla {tanh ( \mathcal{F}(\textbf{w}) )}
= \frac{\partial \mathcal{F}(\textbf{w})}{\partial \textbf{w}} \Big(1 - tanh^2 ( \mathcal{F}(\textbf{w}) ) \Big)
\end{split}
\end{equation}


Assumption 1 derives a corollary that
if $\textbf{w}$ is out of a small range like $[-1,+1]$,
the gradient of $tanh(F())$ for $\textbf{w}$ will not severely vanish.
Through generating the higher dimensional manifolds of full-precision weights $\textbf{w}$,
the gradient vanishing during optimization is relieved,
which allows optimizers to solve the globally optimal.

\paragraph{\textbf{Assumption 2}:} we assume for a vector $v \in \mathbb{R}^n$ that is $k$-sparse,
there exists an extremely small $\epsilon \in (0,1)$ with optimal $\textbf{w}^*_q$,
in the optimization of $\ell(\textbf{w}_q)$ with the objective function $\ell$,
it has the property that

\begin{equation}\label{eq:hypo2}
\begin{split}
&\lim \limits_{\textbf{w}_q \rightarrow \textbf{w}^*_q} \| \ell(\textbf{w}_q) - \ell(sign(\textbf{w}_q)) \|^2_2 = 0 \\
& s.t. \quad (1 - \epsilon) \leq  \frac{\ell(\textbf{w}^*_q v)}{\ell(\textbf{w}^*_q)} \leq  (1 + \epsilon)
\end{split}
\end{equation}

Assumption 2 derives a conclusion that
if the said constraint of $\frac{\ell(\textbf{w}^*_q v)}{\ell(\textbf{w}^*_q)}$ with $\epsilon$ is satisfied,
it represents the top-$k$ elements in $\textbf{w}^*_q$ dominate the objective function $\ell$,
while the remaining elements do not affect the output seriously.
In this case, the discretization error caused by the post-processing step is minimized,
as the sign of top-$k$ elements are equal to themselves.
In brief, the optimization no longer requires all elements in $\textbf{w}_q$ to converge to \{+1,-1\} strictly,
but penalizes it to satisfy the top-$k$ sparse with the task-driven priority.

\subsection{Binarization with the QuantNet}
Based on above two fundamental hypotheses,
we propose a meta-based quantizer named QuantNet for directly generating the binarized weights.
Our proposal is to form the functional $\mathcal{F}$ for transforming $\textbf{w}$
into higher dimensional mainfold $\mathcal{F}(\textbf{w})$,
and optimizing the dominant elements $\textbf{w}_q$ to satisfy the sparse constraint.

As for implementation details of QuantNet,
we design an encoding module accompanied by a decoding module,
and further construct an extra compressing module. 
Specially, suppose full-precision weights come from a convolution layer
with 4D shape $\mathbb{R}^{k \times k \times m \times n}$,
where $k$, $m$ and $n$ denote the kernel size, the number of input channels and the number of output channels, respectively.

The input weights will be firstly reshaped into the 2D shape $\mathbb{R}^{m \cdot n \times k^2}$.
It means that QuantNet is a kernel-wise quantizer and process each kernel of weights independently,
where the batch size is the number of total filters of full-precision weights.
In the encoding and decoding process, it firstly expands the reshaped weights into higher dimensional mainfold,
which is achieved with the dimensional guanrantee that makes
the output shape of encoding module to satisfy $\mathbb{R}^{m \cdot n \times d^2}, \ s.t. \ d \gg k$.
And then, the compressing module is to transform the higher dimensional manifolds into low-dimensional spaces.
If the manifold of interest remains non-zero volume after the compressing process,
it corresponds to a higher priority to improve the performance of binarized model on the specific task.
Finally, the decoding module generates the binarized weights
with the output of the compressing module and the soft binarization function,
while restoring the original shape of full-precision weights for main network optimization.

The Fig. 1 provides visualization of QuantNet in the architecture.

\begin{figure}[t]
  \centering
  \includegraphics[width=0.95\textwidth]{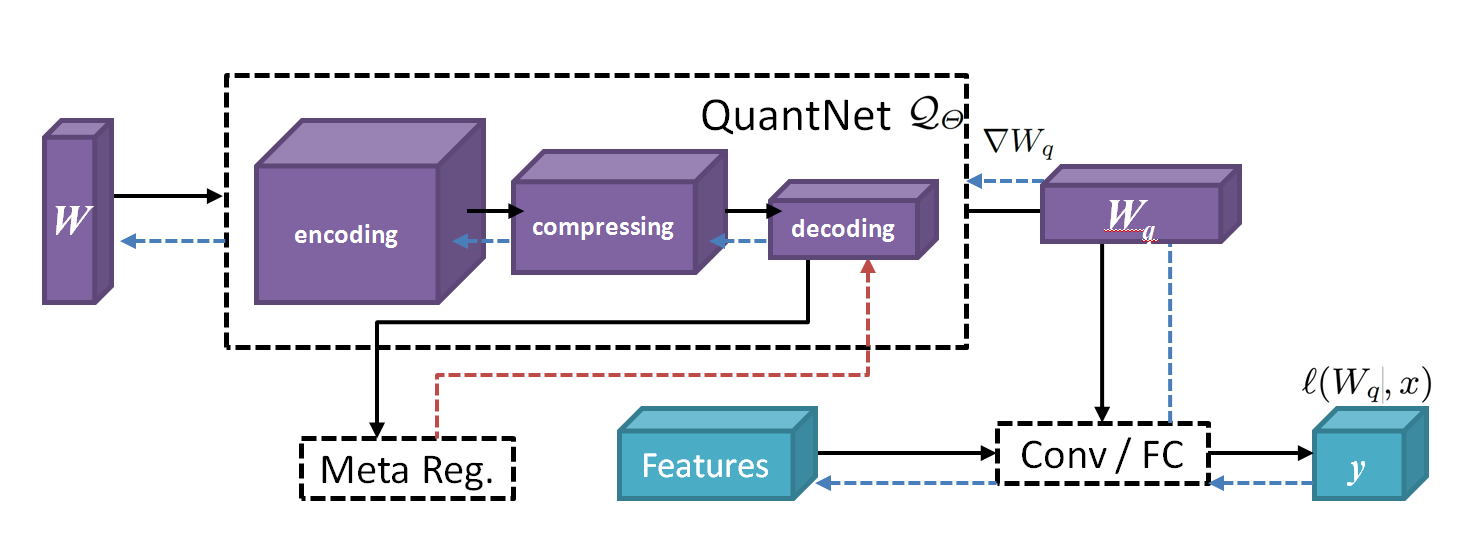}
  \caption{The architecture of QuantNet. QuantNet $\mathcal{Q}_{\Theta}$ takes the full-precision weights $W$ as the input, and directly outputs the binarized weights $W_q$ for the network. With the loss of $\ell(W_q, x)$,  $W_q$ will be directly updated in the back-propagation, and $\nabla W_q$ will be used to update $\Theta$ in QuantNet $\mathcal{Q}$. Finally, a new $W$ is computed during this training step.}\label{fig:QuanNet}
\end{figure}

\paragraph{\textbf{Feed-Forward Step}}
Given a full-precision weights\footnote{We omit the notation of layers $l$ in $W_l$ for simplification.} $W$,
the proposed QuantNet $\mathcal{Q}_{\Theta}$ incorporates the parameters $\Theta$
to generate the binarized weights $W_q$ with $tanh(\mathcal{Q}_{\Theta}(W))$.
After $W$ is quantized as $W_q$, the loss $\ell$ is generated by $\ell(W_q, \{x, y\})$ with the training set \{x, y\}.

\paragraph{\textbf{Back-Propagation Step}}
The gradient of $\ell$ with regard to $W_q$ in each layer is computed by the back-propagation.
For example, the gradient of weights $g_{W_q}$ in last layer is computed by $\nabla \ell (W_q)$.
Then, the QuantNet $\mathcal{Q}_{\Theta}$ receives the $g_{W_q}$ from the corresponding $W_q$,
and updates its parameters $\Theta$ by

\begin{equation}\label{eq:update_quantnet}
g_{\Theta} = \frac{\partial \ell}{\partial W_q} \frac{\partial W_q}{\partial \Theta} = g_{W_q} \frac{\partial W_q}{\partial \Theta}
\end{equation}

and the gradients of $g_{\Theta}$ is further used to update the full-precision weights $W$ by,

\begin{equation}\label{eq:update_fp}
W^{t+1} = W^{t} - \eta \cdot g_{\Theta}
\end{equation}

where $t$ denotes the $t$-th training iteration and $\eta$ is the learning rates defined for QuantNet.

In practice, QuantNet is applied layer-wise.
However, as the number of extra parameters introduced by QuanNet is much less than the network weights,
so the computation cost caused by our proposal is acceptable during the network training as shown in the Tab. \ref{tab:complex}.

\subsection{Optimization}
As for the optimization of QuantNet,
it is included in the target network that will be binarized.
Given the full-precision weights, QuantNet generates the binarized weights to apply on objective tasks,
and it is optimized with the back-propagation algorithm to update all variables.
In brief, our binarization framework is fully differentiable without any gradient estimators,
so there is no information loss during the binarization.
We now present the optimization details.

\begin{algorithm}[t] \label{alg:op}
  \caption{Generating the binarized weights with QuantNet}
  \KwIn{the full-precision weights $W$, the QuantNet $\mathcal{Q}$ with parameters $\Theta$, and the training set \{X, Y\}, training iteration $t$, $\epsilon = 1e-5$}
  \KwOut{the optimally binarized weights $W^*_q$}
  \textbf{Training for each layer}\\
  \For{$t=0;t \le T$}
  {
    $\textbf{Feed-Forward}$\;
    \quad Compute $W^t_q$ with $tanh(\mathcal{Q}_{\Theta^t}(W^t))$\;
    \quad Compute $\ell(W^t_q, \{x^t,y^t\})$ with $W^t_q$ and $\{x^t, y^t\}$\;
    $\textbf{Back-Propagation}$\;
    \quad Compute $\nabla W^t_q$ with $\ell(W^t_q, \{x^t, y^t\})$\;
    \quad Compute $\nabla \Theta^t$ with Eq. \ref{eq:update_quantnet} and Eq. \ref{eq:quan_op}\;
    \quad Update the $W^t$ with Eq. \ref{eq:update_fp}\;
  }
  \textbf{Discretization Step}\\
  $W^*_q = sign(W^T_q)$\;
\end{algorithm}

\paragraph{\textbf{QuantNet Optimization}}
For satisfying the constraint in assumption 2, we propose an objective function inspired by the idea of sparse coding.
It constrains the compressing process in QuantNet during the binarization.
Let $\textbf{w}_q$ be the binarized weights and introduce a reference tensor $\textbf{b}$,
we aim to find an optimal $\textbf{w}^*_q$ that satisfies

\begin{equation}\label{eq:quan_op}
\textbf{w}^*_q = \arg \min \limits_{\textbf{w}_q} \| \textbf{b} - \sqrt{\textbf{w}^2_q} \|_2
+ \| \textbf{w}_q \|_1, \ s.t. \ \textbf{b} \in \{1\}^{m \cdot n}
\end{equation}
where, tensor $\textbf{b}$ is chosen to be all ones to make elements in $\textbf{w}_q$ to get close to -1 or +1.
As Eq. \ref{eq:quan_op} is independent of the task optimization of binarized models,
we alternately solve it with an extra optimizer during the standard network training.
At each iteration of optimization, there is adversarial relationship
between Eq. \ref{eq:update_quantnet} and Eq. \ref{eq:quan_op},
and the optimization tries to find the balance between minimizing the binarization error
while penalizing the sparsity of binarized weights based on the task priority.

\paragraph{\textbf{Binarized Model Optimization}}
As for the optimization of binarized model,
we use the standard mini-batch based gradient descent method.
After the QuantNet $\mathcal{Q}_{\Theta}$ is constructed and initialized,
the QuantNet optimization is accompanied with the training process of the binarized model,
and the objective function of binarized model optimization depends on the specific task.
With the specific objective function,
the task-driven optimizer is employed to compute the gradients for each binarized layer.
The operations for whole optimization are summarized in Algorithm \ref{alg:op}.

\section{Experiments}
In this section, we firstly show the implementation details of our method and experiment settings.
Secondly, the performance comparison between our method and STE-based or Non STE-based methods is generated,
which further includes the analysis of convergence behaviour.

\subsection{Implementation Details}
As for the implementation details in experiments,
we run each experiment five times with the same initialization function from different starting points.
Besides, we fix the number of epoches for training and
use the same decay strategy of learning rate in all control groups.
At the end, we exhibit the average case of training loss and corresponding prediction accuracy.
We show the implementation details as follows.

\paragraph{\textbf{Network Architecture}}
We apply the unit with structure of ``FC-BN-Leaky ReLU'' to construct QuantNet,
and each processing module in QuantNet contains at least one unit.
For reducing the complexity of network training,
QuantNet used in experiments contains only one unit for each processing module,
and we still observe a satisfied performance during evaluation.
Similar to the existing methods \cite{rastegari2016xnor,Zhang2018LQ,Uhlich2019DQ},
we leave the first and last layers and then binarizing the remaining layers.
For comparison, a fully binarized model by binarizing all layers is also generated.
Considering that the bitwise operations can speedup the inference of network significantly,
we analyze the balance between the computation cost saving and model performance boosting by these two models.
The experiment result exhibits only 0.6\% accuracy drop (more than 1 \% in previous methods)
in CIFAR-10 \cite{cifar10} with ResNet-20, in the case that all layers are binairzed.

\paragraph{\textbf{Initialization}}
In experiments, all compared methods including our method
use the truncated Gaussian initialization if there is not specified in their papers,
and all binarized model from experiments are trained from scratch without leveraging any pre-trained model.
As for the initialization of QuantNet, we employ the normal Gaussian initialization for each layer.
Furthermore, we also evaluate the random initialization, which
initialize the variable with the different settings of the mean and variance,
but there is not significant difference on the results.

\paragraph{\textbf{Hyper-parameters and Tuning}}

We follow the hyper-parameter settings such as the learning rate, batch size,
training epoch and weight decay of their original paper.
For fair comparison, we use the default hyper-parameters
in Meta-Quant \cite{chen2019meta} and Self-Binarizing \cite{lahoud2019self} to generate the fully binarized network.
As for the hyper-parameters of our QuantNet,
we set the learning rate as $1e-3$ and the moving decay factor as $0.9$.
We also evaluate different optimization methods including
SGD(M) \cite{sgd1951}, Adam \cite{adam2014} and AMSGrad \cite{amsgrad2018}.
Although we observe that the soft binarization in AMSGrad
has a faster convergence behaviour than the others,
we still use the SGD(M) with average performance for all methods to implement the final comparison.
In future, we plan to analyze the relationship between the soft binarization and different optimizers.
\begin{table}[htb]
\begin{center}
	\centering
    \begin{tabular}{ccc}
    \hline
    Optimizer       & Accuracy    & Training   \\
                    &  (\%)       &  time      \\
    \hline
    SGD(M)         & 90.04     &   1.0 $\times$     \\
    Adam           & 89.98     &   \textasciitilde 1.2 $\times$     \\
    AMSGrad        & 90.12     &   \textasciitilde 0.9 $\times$     \\
    \hline
    \noalign{\smallskip}
    \end{tabular}
    \caption{Comparison with different optimizer on ResNet-20 for CIFAR10.}
    \label{tab:perf_op}
\end{center}
\end{table}

\subsection{Performance Comparison}
QuantNet aims at generating the binarized weights without STE and other estimators.
Hence, we compare it with both STE-based binarization
methods \cite{courbariaux2015binaryconnect,Zhou2016DoReFa,hou2016loss,Zhang2018LQ,jung2019learning}
and Non STE-based binarization
methods \cite{bai2019prox,chen2019meta,Uhlich2019DQ,lahoud2019self}
with the idea of avoiding the discrete quantization.
In details, the evaluation is based on the standard benchmark of classification task
with CIFAR-100 \cite{cifar10} and ImageNet \cite{Deng2009},
and the base network architectures are based on the
AlexNet \cite{Krizhevsky2012}, ResNet \cite{he2016deep} and MobileNet \cite{sandler2018mobilenetv2}.

\paragraph{\textbf{Evaluation of the Discretization Error}}
As the soft binarization method \cite{lahoud2019self} always involves a post-processing step,
which aims at transforming the float-point weights into the fixed-point weights,
we name this step the discretization step which is shown in Algorithm \ref{alg:op}.
For comparing the discretization error caused by the step
between the self-binarizing \cite{lahoud2019self} and the proposed QuantNet,
we generate both two binarized models for self-binarizing and QuantNet,
and use the notation (D) to denote the prediction accuracy of binarized model after the discretization step,
which means the weights in the binarized model is transformed into the integer exactly.
As shown in the Tab. \ref{tab:perf_drop}, QuantNet achieves the best performance even better than the FP,
the major reason is that the sparse constraint encourages a better generalization ability.
Moreover, since the discretization error is considered in our algorithm during the binarizing process,
comparing to the accuracy drop 1.85\% in self-binarizing \cite{lahoud2019self},
QuantNet only reduces 0.59\%.

\begin{table}[htb]
\begin{center}
   \centering
   \begin{tabular}{cccc}
   \hline
    Methods                       & Bit-width (W/A)   & Acc.(\%)             & Acc.(\%) (after discretization) \\
   \hline
    FP                                    & 32/32     &  86.55               & - \\
    Self-Binar. \cite{lahoud2019self}     & 1/1       &  86.91\%             & 84.31\%  \\
    Ours                                  & 1/1       &  \textbf{87.08} \%   & \textbf{86.49} \%    \\
    \hline
    \noalign{\smallskip}
    \noalign{\smallskip}
    \end{tabular}
    \caption{Prediction accuracy of binarized AlexNet on CIFAR-10. The FP represents the full-precision model with 32-bits for both weights and activations.}
    \label{tab:perf_drop}
\end{center}
\end{table}

\paragraph{\textbf{Comparison with STE-based Binarization}}
We evaluate our QuantNet with the STE-based binarization methods,
and report the top-1 accuracy in Tab. \ref{tab:comp_cifar}.
Besides, we use PACT \cite{pact2018} to quantize the activation into 2 bits
if the compared method does not support the activation quantization.
For the compared prediction accuracy used in this table,
we use the results from the original paper if it is specified.
Overall, QuantNet achieves the best performance compared to existing STE-based methods,
which surpasses QIL \cite{jung2019learning} more than 2\% before the discretization step,
and even obtain a comparable performance with the full-precision model.
It demonstrates the advantage of directly binarizing the weights within a fully differentiable framework.
Although the discretization (rounding operation) introduces a post-processing step,
the experiment results still prove the effectiveness of our binarization method,
and the degradation of prediction accuracy caused by rounding is negligible.

\begin{table}[htb]
\begin{center}
    \begin{tabular}{cccc}
    \hline
    Method & Bit-width & ResNet-56 & ResNet-110    \\
    \hline
    FP                                      &32W32A & 71.22 \% & 72.54 \%   \\
    BWN \cite{courbariaux2015binaryconnect} &1W2A   & 64.29 \% & 66.26 \%   \\
    DoReFa-Net \cite{Zhou2016DoReFa}        &1W2A   & 66.42 \% & 66.83 \%   \\
    LAB \cite{hou2016loss}                  &1W2A   & 66.73 \% & 67.01 \%   \\
    LQ-Nets \cite{Zhang2018LQ}              &1W2A   & 66.55 \% & 67.09 \%   \\
    QIL \cite{jung2019learning}             &1W2A   & 67.23 \% & 68.35 \%   \\
    Ours                                    &1W2A   & \textbf{69.38} \% & \textbf{70.17} \%  \\
    Ours(D)                                 &1W2A   & \textbf{68.79} \% & \textbf{69.48} \%  \\
    \hline
    \noalign{\smallskip}
    \end{tabular}
    \caption{Top-1 accuracy (\%) on CIFAR-100. Comparison with the existing methods on ResNet-56 and ResNet-110.}
    \label{tab:comp_cifar}
\end{center}
\end{table}

\paragraph{\textbf{Comparison with Non STE-based Binarization}}
As for the Non STE-based binarization methods,
it mainly includes two categories: \emph{learning better gradients for non-differentiable binarization function},
and \emph{replacing the non-differentiable function with the differentiable one}.
We compare the QuantNet with the representative works in these two categories -
ProxQuant \cite{bai2019prox} and Meta-Quant \cite{chen2019meta} in the first
and Self-Binarizing \cite{lahoud2019self} in the second.
With the increasing number of parameters in larger architecture \cite{he2016deep},
although the methods \cite{bai2019prox,chen2019meta}
related gradient refinement have improved the performance effectively,
the bottleneck caused by the gradient estimation appears obviously,
and our method have achieved the significant improvement than these methods (Tab. \ref{tab:comp_imagenet}).
Moreover, as the discretization error caused by the rounding operation is well considered by our method,
QuantNet is affected less than Self-Binarizing \cite{lahoud2019self}.

\begin{table}[htb]
\begin{minipage}{0.48\linewidth}
\centering
\begin{tabular}{ccc}
\hline
Method & Bit-width & AlexNet\\
\hline
FP                                        & 32W32A & 55.07 \% \\
Self-Binar.  \cite{lahoud2019self}        & 1W32A  & 52.89 \% \\
Self-Binar.(D) \cite{lahoud2019self}      & 1W32A  & 50.51 \%\\
Ours                                      & 1W32A  & \textbf{54.06} \% \\
Ours(D)                                   & 1W32A  & \textbf{53.59} \% \\
\hline
\centering
\end{tabular}
\end{minipage}\begin{minipage}{0.48\linewidth}
\centering
\begin{tabular}{ccc}

\hline
Method & Bit-width & ResNet-34 \\
\hline
FP \cite{Qin2020}                         & 32W32A & 73.30 \%  \\
ProxQuant  \cite{bai2019prox}             & 1W32A  & 70.42 \%   \\
Meta-Quant \cite{chen2019meta}            & 1W32A  & 70.84 \%   \\
Ours                                      & 1W32A  & \textbf{71.97} \% \\
Ours(D)                                   & 1W32A  & \textbf{71.35} \% \\
\hline
\centering
\end{tabular}
\end{minipage}
\caption{Top-1 accuracy (\%) on ImageNet. Comparison with the existing methods on AlexNet (left), ResNet-34 (right).}
\label{tab:comp_imagenet}
\end{table}

\paragraph{\textbf{Convergence Analysis}}
We analyze the convergence behaviour of our QuantNet and other binarization methods during the training process.
In details, we use ResNet-34 as the base architecture,
and compare with the STE-based methods and non-STE based methods separately.
For the first case, 
QuantNet exhibits a significantly smooth loss curve over STE,
including much faster convergence speed and lower loss values,
and it also achieves the best prediction accuracy in the test reported in Tab. \ref{tab:comp_cifar}.
The main reason of the better convergence of our method is that
QuantNet is totally differentiable during the optimization.
Furthermore, we analyze the proposed method in the second case,
and we observe that all the non-STE based methods can smooth the loss curve effectively,
but our method achieve the lowest loss value as there is not estimation of gradients.


\paragraph{\textbf{Complexity of Models}}
\begin{wraptable}{r}{6.0cm}
	\centering
    \begin{tabular}{cc}
    \hline
    Bit-width          &    Training        \\
    (1W/2A)            &  time(iter./s)     \\
    \hline
    FP(32W/32A)                          & 1.0  $\times$     \\
    MixedQuant   \cite{Uhlich2019DQ}     & 1.5  $\times$    \\
    Meta-Quant   \cite{chen2019meta}     & 2.8  $\times$    \\
    Self-Binar.  \cite{lahoud2019self}   & 1.2  $\times$    \\
    Ours                                 & 1.7  $\times$    \\
    \hline
	\end{tabular}
    \caption{Training time on ResNet-34}
    \label{tab:complex}
\end{wraptable}

As our QuantNet involves the extra computation cost and parameters during the optimization,
we analyze its efficiency comparing to the traditional STE-based methods and other Meta-based methods.
For QuantNet, it is independent of the scale of input resource,
and its time complexity is related to the amount of its parameters.
In the Tab. \ref{tab:complex}, the total training time cost is exhibited,
and we leave the inference step since the QuantNet is removed in this step.
For the setting of experiment in this table, the base architecture ResNet-34 is used,
and the bitwise operation is not implemented for all cases.

\section{Conclusions}
In the paper, we present a meta-based quantizer QuantNet to binarize the neural network,
which directly binarize the full-precision weights without STE and any learnable gradient estimators.
In contrast to the previous soft binarizing method,
the proposed QuantNet not only solves the problem of gradient vanishing during the optimization,
but also alleviates the discretization errors caused by the post-processing step for obtaining the fixed-point weights.
The core idea of our algorithm is to transform the high dimensional manifolds of weights,
while penalize the dominant elements in weights into sparse according to the task-driven priorty.
In conclusion, the QuantNet outperforms the existing binarization methods on the standard benchmarks,
which not only can be applied on weights, but also can be extended to activations (or quantization with other bits) easily.

%
%
\bibliographystyle{splncs04}
\bibliography{egbib}
\end{document}